\documentclass[sigconf]{acmart}

\copyrightyear{2026}
\acmYear{2026}
\setcopyright{cc}
\setcctype{by}
\acmConference[KDD 2026] {Proceedings of the 32nd ACM SIGKDD Conference on Knowledge Discovery and Data Mining V.2}{August 9--13, 2026}{Jeju Island, Republic of Korea.}
\acmBooktitle{Proceedings of the 32nd ACM SIGKDD Conference on Knowledge Discovery and Data Mining V.2 (KDD 2026), August 9--13, 2026, Jeju Island, Republic of Korea}
\acmISBN{979-8-4007-2259-2/2026/08}
\acmDOI{10.1145/3770855.3818430}

\usepackage{tabularx} 
\usepackage{makecell}
\usepackage{threeparttable}
\usepackage{booktabs} 
\usepackage{multirow} 
\usepackage{array}    
\usepackage{graphicx} 
\usepackage{subcaption} 
\usepackage{enumitem}

\begin{document}

\title{Hi-SAM: A Hierarchical Structure-Aware Multi-modal Framework for Large-Scale Recommendation}
\author{Pingjun Pan}
\authornote{Both authors contributed equally to this research.}
\email{panpingjun@corp.netease.com}
\orcid{0009-0007-9892-9772}
\affiliation{%
  \institution{Netease Cloud Music}
  \city{Hangzhou}
  \country{China}
}

\author{Tingting Zhou}
\authornotemark[1]
\email{hzzhoutingting15@corp.netease.com}
\affiliation{%
  \institution{Netease Cloud Music}
  \city{Hangzhou}
  \country{China}
}

\author{Peiyao Lu}
\email{lupeiyao@corp.netease.com}
\affiliation{%
  \institution{Netease Cloud Music}
  \city{Hangzhou}
  \country{China}
}

\author{Tingting Fei}
\email{feitingting@corp.netease.com}
\affiliation{%
  \institution{Netease Cloud Music}
  \city{Hangzhou}
  \country{China}
}

\author{Hongxiang Chen}
\authornote{Hongxiang Chen is the corresponding author.}
\email{hzchenhongxiang@corp.netease.com}
\affiliation{%
  \institution{Netease Cloud Music}
  \city{Hangzhou}
  \country{China}
}

\author{Chuanjiang Luo}
\email{luochuanjiang03@corp.netease.com}
\affiliation{%
  \institution{Netease Cloud Music}
  \city{Hangzhou}
  \country{China}
}
\renewcommand{\shortauthors}{Pingjun Pan et al.}

\begin{abstract}
In recent years, multi-modal recommendation has attracted increasing attention, as items inherently possess rich semantic attributes such as text descriptions and cover images. Semantic ID-based approaches have demonstrated effectiveness by discretizing multi-modal information into compact discrete token representations. However, two critical challenges persist: (1) \textbf{Suboptimal Multi-modal Tokenization}: existing quantization methods (e.g., RQ-VAE) lack explicit disentanglement between shared cross-modal semantics and modality-specific details, causing information redundancy or modality collapse; (2) \textbf{Architecture-Data Mismatch}: vanilla Transformer architectures treat semantic ID sequences as flat token streams, ignoring the intrinsic hierarchy spanning user interactions, item sequences, and fine-grained tokens. Moreover, expanding each item into multiple tokens amplifies sequence length and accumulates noise, biasing attention toward local details while neglecting holistic item semantics.

To address these challenges, we propose Hi-SAM, a Hierarchical Structure-Aware Multi-modal framework with two key designs: (1) \textbf{Disentangled Semantic Tokenizer (DST)}, which unifies heterogeneous modalities via geometry-aware alignment on a shared hypersphere, and quantizes them through a coarse-to-fine strategy—shared codebooks distill cross-modal consensus while modality-specific codebooks recover complementary nuances from residuals, enforced by mutual information minimization to ensure explicit disentanglement; (2) \textbf{Hierarchical Memory-Anchor Transformer (HMAT)}, which splits positional encoding into inter-item and intra-item orthogonal subspaces via Hierarchical RoPE to restore the flattened hierarchy, and inserts Anchor Tokens that condense each item into a compact memory—retaining fine-grained details for the current item while accessing historical items only through their compressed summaries. Extensive experiments and ablation studies on real-world datasets demonstrate consistent improvements over state-of-the-art baselines, especially in cold-start scenarios. Hi-SAM has been deployed on a large-scale social platform serving millions of daily users, achieving a \textbf{6.55\%} gain in the core online business metric.

\end{abstract}


\maketitle

\section{Introduction}
In recent years, the paradigm of recommender systems has been profoundly reshaped by Large Model architectures. Inspired by the success of Transformers in natural language processing, prior research~\cite{kaplan2020scaling, ardalani2022understanding, zhang2024scaling, shin2023scaling} has demonstrated that scaling up model parameters and training data yields significant performance gains in recommendation tasks. Prominent sparse ID-based large models, such as \cite{zhang2024wukong, zhai2024actions}, have validated this scaling law. However, these methods are fundamentally constrained by their excessive reliance on Sparse IDs. While \cite{han2025mtgr} have sought to mitigate this by incorporating cross-features (e.g., CTR), these auxiliary signals are essentially statistical aggregations derived from ID-based interactions rather than intrinsic content representations. Consequently, such approaches remain highly susceptible to performance degradation in cold-start scenarios where interaction data is scarce. Crucially, they fail to leverage the rich multi-modal semantics (e.g., visual appearance, textual descriptions) inherent to items. These multi-modal attributes provide a comprehensive depiction of item utility and hold significant potential for enhancing recommendation accuracy~\cite{huang2019multimodal,mu2023multimodal}, yet remain overlooked by ID-based paradigms.

Recent studies have explored Semantic ID-based recommenders~\cite{rajput2023recommender,singh2024better,luo2025qarm}. This paradigm hinges on two critical modules: Semantic ID Generation, which maps similar items to shared discrete codes to enhance generalization, and Large Recommendation Model Architecture, which leverages large transformer-based models for prediction. For Semantic ID Generation, independent quantization methods~\cite{wang2025progressive,qiao2026text} process each modality separately, causing redundancy from overlapping semantics (e.g., visual ``vintage jacket'' vs. textual ``retro coat'') and fragmented representations that miss cross-modal interactions. Fusion-based methods\cite{luo2025qarm, zheng2025personalized} integrate modalities before quantization (e.g., QARM~\cite{luo2025qarm} trains unified encoders for early fusion), but such indiscriminate mixing often leads to \textit{modality collapse}~\cite{peng2022balanced}, where dominant modalities overshadow critical details from others. Regarding Large Model Architecture, transforming user behavior sequences into semantic ID sequences flattens the item-level hierarchy, since each item becomes multiple tokens. This introduces two issues: (1) cross-item and within-item token transitions become indistinguishable (e.g., adjacent tokens across items have distance 1), obscuring item boundaries; (2) models may over-focus on fine-grained attribute tokens while missing holistic item semantics. Contemporary Transformer backbones (e.g., Qwen~\cite{bai2023qwen}, HSTU~\cite{zhai2024actions}) are designed for flat sequences and inherently overlook this hierarchical structure.

To address these challenges, we propose a core insight: \textit{better discrete semantic IDs combined with better-adapted model architecture yield superior recommendation performance}. Based on this insight, we identify the need for systematic improvements at two levels: (1) Semantic ID Generation. An effective semantic ID system must capture rich multimodal item information while maintaining lightweight generation process; (2) Model Architecture. The architecture must be tailored to the structured nature of semantic IDs in recommendation scenarios, effectively balancing the utilization of coarse- and fine-grained information without incurring additional computational overhead. To this end, we propose \textbf{Hierarchical Structure-Aware Multimodal Framework (Hi-SAM)}. Hi-SAM adopts a two-stage architecture: the first stage employs a \textbf{Disentangled Semantic Tokenizer (DST)} to map multimodal item content into high-quality discrete semantic IDs; the second stage leverages a \textbf{Hierarchical Memory-Anchor Transformer (HMAT)} to perform hierarchical sequence modeling and preference prediction based on these semantic IDs.

In the DST module, we adopt the fusion-based method. We first employ Gramian Representation Alignment Measure to project representations from different modalities into a higher-dimensional space and perform geometric alignment by minimizing the volume of the parallelotope spanned by multimodal vectors, ensuring alignment of different modalities within a unified semantic space through a lightweight approach~\cite{cicchetti2025gramian}. Subsequently, we propose Disentangled Modal-Residual Quantization to quantize the aligned multimodal representations, which employs a coarse-to-fine quantization strategy. The shared layers capture cross-modal commonalities through residual quantization to avoid information redundancy, while the modality-specific layers leverage semantic-guided attention mechanisms to recover modality-specific details from residuals, preventing modality collapse during multimodal fusion. An explicit mutual information constraint enforces disentanglement between shared and specific representations. This approach enables the generated semantic IDs to more comprehensively express item attributes.

In the HMAT module, we explicitly account for the hierarchical structure of recommendation data and propose two tailored adaptations. First, we introduce Hierarchical RoPE, which decouples the positional encoding space into two orthogonal subspaces: inter-item positions with larger base frequencies for long-range dependency modeling, and intra-item positions with smaller base frequencies for fine-grained local sensitivity. Second, we propose Memory-Anchor Attention, which inserts a special Anchor Token after each item to serve as a compressed semantic summary. Through structured masking, the model attends to all tokens within the current item for fine-grained information extraction, while restricting interactions with historical items exclusively to their Anchor Tokens. This integration into Transformer attention yields two key advantages:(1) it reduces noise propagation from token-level variations across long sequences, improving model robustness; (2) it substantially reduces the attention complexity incurred by expanding each item into multiple tokens, while maintaining expressive power through the compressed anchor representations. Additionally, we employ a two-stage progressive training strategy that decouples semantic representation learning from preference modeling via unsupervised semantic pretraining followed by supervised fine-tuning on recommendation objectives. Our main contributions are as follows:

\begin{itemize}[leftmargin=*]
 \item We propose \textbf{Hi-SAM}, a novel hierarchical structure-aware multi-modal framework addressing the tokenization–architecture gap in semantic ID-based recommendation, comprising a Disentangled Semantic Tokenizer and a Hierarchical Memory-Anchor Transformer.

    \item In DST, we design a geometry-aware Cross-Modal Alignment and a novel Disentangled Modal-Residual Quantization to decouple cross-modal consensus from modality-specific nuances. In HMAT, we propose Hierarchical RoPE to restore the flattened item–attribute hierarchy, and a biologically-inspired Memory-Anchor Attention that condenses history into compact memories to mitigate noise.
    
  \item Extensive offline and online experiments validate Hi-SAM's superiority, with a \textbf{6.55\%} lift in the core business metric and \textbf{35\%} lower latency in production.
\end{itemize}

\section{Related works}
Multimodal information has been progressively integrated into recommender systems to complement sparse collaborative signals. Early DLRMs incorporated multimodal features as side information, from CNN visual features~\cite{he2016vbpr} to graph-based latent structures~\cite{zhang2021mining, zhang2022latent}. Recent approaches leverage pre-trained encoders such as CLIP~\cite{radford2021learning} and Sentence-BERT~\cite{reimers2019sentence} for higher-quality representations~\cite{zhou2023comprehensive, yuan2023go}, with aggregation strategies such as feature concatenation~\cite{ngiam2011multimodal}, independent encoding~\cite{gadzicki2020early}, cross-attention~\cite{wei2023multi}, and gating mechanisms~\cite{ma2018entire}. Beyond continuous representations, the Semantic ID paradigm discretizes item representations into compact token sequences via vector quantization~\cite{lee2022autoregressive, rajput2023recommender, hou2023learning, hou2022towards, singh2024better, luo2025qarm}.

The evolution of recommendation architectures has progressed from shallow models to deep architectures and more recently toward large-scale Transformer-based frameworks. Early deep models such as Wide \& Deep~\cite{cheng2016wide}, DeepFM~\cite{guo2017deepfm}, and DCN~\cite{wang2017deep, wang2021dcn} combined feature interaction modules with deep networks but operated without sequential modeling. The introduction of attention mechanisms catalyzed a shift toward sequence-aware architectures: DIN~\cite{zhou2018deep} employed target-aware attention for adaptive behavior aggregation, while DIEN~\cite{zhou2019deep} captured evolving user interests through interest evolution networks. Transformer-based architectures subsequently became the dominant backbone, with SASRec~\cite{kang2018self} adapting unidirectional Transformers for next-item prediction and BERT4Rec~\cite{sun2019bert4rec} introducing bidirectional self-attention with masked item prediction. At industrial scale, HSTU~\cite{zhai2024actions} proposed pointwise aggregated attention tailored for user action sequences, and Wukong~\cite{zhang2024wukong} validated the scaling law in recommendation with stacked factorization machines. Research at the intersection of LLMs and recommendation~\cite{wu2024survey, bao2023tallrec, geng2022recommendation} has further explored leveraging pre-trained language models through prompt-based methods or generative formulations.

\section{Methodology}

\subsection{Formulation and Framework}
\noindent  \textbf{Problem Formulation.} Let $\mathcal{U}$ and $\mathcal{I}$ denote the set of users and items, respectively. For any user $u \in \mathcal{U}$, the historical interaction sequence is ordered chronologically. We define the behavior item sequence as $S_{u,i} = \{i_1, i_2, \dots, i_k\}$ and the corresponding action sequence as $S_{u,a} = \{a_1, a_2, \dots, a_k\}$, where $a_t \in \mathcal{A}$ represents the interaction type (e.g., click, reply) and $i_t \in \mathcal{I}$ denotes the interacted item at step $t$, respectively, and $k$ is the sequence length. For each item $i \in \mathcal{I}$, we define its raw multi-modal feature set as $\mathcal{X}_i = \{x_{i,1}, x_{i,2}, \dots, x_{i,N_{m}}\}$, where $N_{m}$ is the number of modalities, and $x_{i,j}$ denotes the raw data of the $j$-th modality (e.g., image, text). Consequently, the user's history can be represented in multi-modal form as $S_{u,m} = \{\mathcal{X}_{i_1}, \mathcal{X}_{i_2}, \dots, \mathcal{X}_{i_k}\}$. The goal of our proposed multi-modal recommendation framework is to predict the probability of user $u$ performing action $a_{k+1}$ on a target item $i_{k+1}$. Formally, we estimate $P(a_{k+1} \mid S_{u,m}, S_{u,a}, \mathcal{X}_{k+1})$.

\noindent  \textbf{Framework Overview.} As illustrated in Figure 1, our Hi-SAM framework consists of two stages: Disentangled Semantic Tokenizer (DST) and Hierarchical Memory-Anchor Transformer (HMAT). In the DST stage, we generate discrete semantic IDs from the raw multi-modal features $\mathcal{X}_i$ of each item. In the HMAT stage, we encode the user's item sequence into semantic token sequence using these discrete IDs, and model user interests through the hierarchical attention mechanism.


\begin{figure*}[t]
  \centering
  \includegraphics[width=1.0\linewidth]{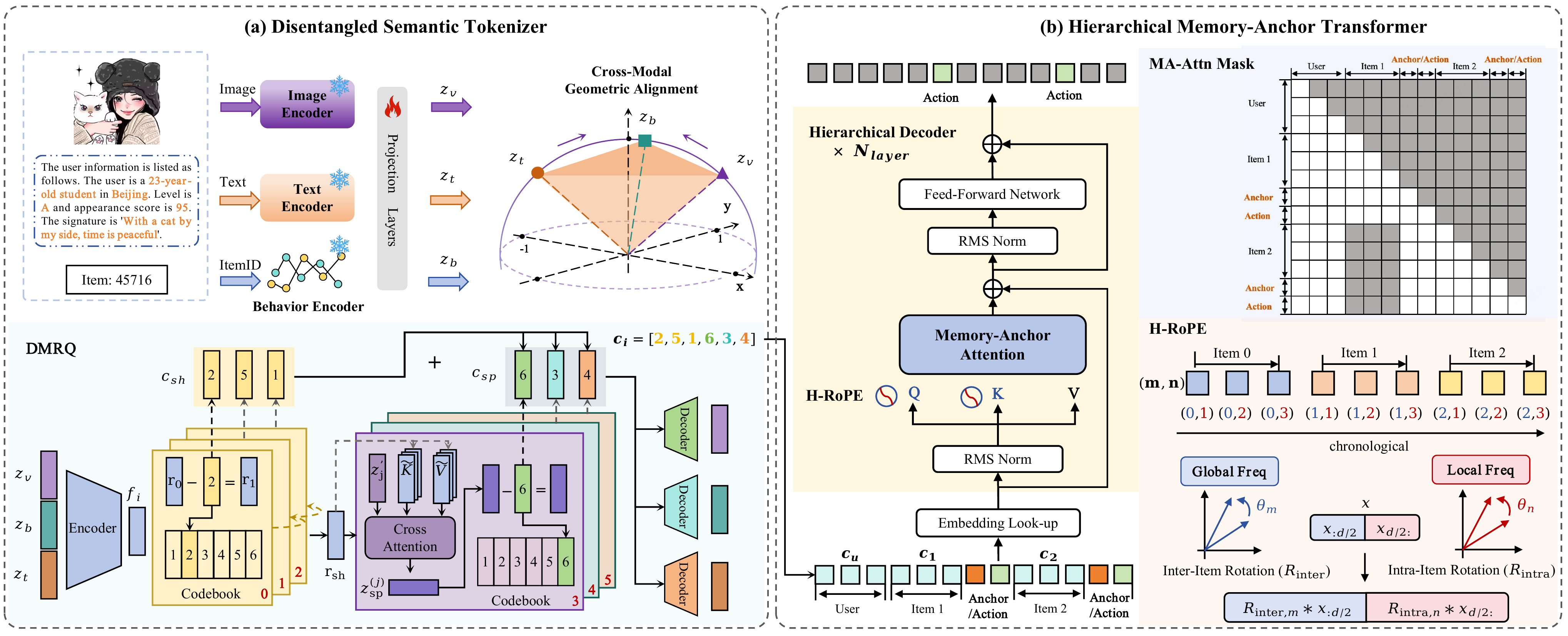}
  \caption{The architecture of Hi-SAM, which consists of the Disentangled Semantic Tokenizer (DST) stage and the Hierarchical Memory-Anchor Transformer (HMAT) stage.}
  \Description{Hi-SAM framework}
  \label{Hi-SAM}
\end{figure*}

\subsection{Disentangled Semantic Tokenizer}
\label{dst}

As illustrated in~\ref{Hi-SAM}(a), DST consists of two modules: (1) \textbf{Cross-Modal Geometric Alignment (CGA)}, which unifies modalities on a hypersphere, and (2) \textbf{Disentangled Modal-Residual Quantization (DMRQ)}, which encodes them by decoupling shared consensus from specific nuances.

\subsubsection{\textbf{Cross-Modal Geometric Alignment (CGA)}}
Conventional multi-modal alignment methods rely on pairwise alignment (e.g., CLIP) to align $N_{m} > 2$ modalities. However, they lack a holistic center, often leading to \textit{subspace fragmentation}, where embeddings of different modalities for the same item remain distinct space even after alignment~\cite{li2025vt}. To address this, we adopt the GRAM \citep{cicchetti2025gramian}, which aligns all modalities simultaneously by minimizing the volume of the parallelotope spanned by their embeddings.


For each modality $j$, we use a specific encoder $E_{\phi_j}$ and projection head $W_j$ to map raw data $x_{i,j}$ to a common dimension $d$. Crucially, we strictly normalize the embeddings to the unit hypersphere, $\mathbf{z}_{i,j} = \frac{W_j E_{\phi_j}(x_{i,j})}{\| W_j E_{\phi_j}(x_{i,j}) \|_2}$, to prevent geometric collapse. We then construct the Gram matrix $\mathbf{G}_i \in \mathbb{R}^{N_{m} \times N_{m}}$ where $(\mathbf{G}_i)_{j,k} = \mathbf{z}_{i,j}^\top \mathbf{z}_{i,k}$. The geometric coherence is quantified by the volume $\text{Vol}_i = \sqrt{\det(\mathbf{G}_i)}$. A smaller volume indicates that the multi-modal vectors are tightly clustered, effectively mitigating subspace fragmentation.

To learn this structure, we designate one modality as the anchor $\mathbf{a}_i$ and the rest as data $\mathbf{r}_i$. We employ a symmetric contrastive loss to minimize the volume for matched pairs while maximizing it for mismatched ones:
\begin{equation}
\begin{aligned}
    \mathcal{L}_{D2A} &= - \frac{1}{B} \sum_{i=1}^{B} \log \frac{e^{-\text{Vol}(\mathbf{a}_i, \mathbf{r}_i)/\tau}}{\sum_{k=1}^{K} e^{-\text{Vol}(\mathbf{a}_k, \mathbf{r}_i)/\tau}}, \\
    \mathcal{L}_{A2D} &= - \frac{1}{B} \sum_{i=1}^{B} \log \frac{e^{-\text{Vol}(\mathbf{a}_i, \mathbf{r}_i)/\tau}}{\sum_{k=1}^{K} e^{-\text{Vol}(\mathbf{a}_i, \mathbf{r}_k)/\tau}}
\end{aligned}
\end{equation}
The total alignment loss is $\mathcal{L}_{align} = (\mathcal{L}_{D2A} + \mathcal{L}_{A2D})/2$. This ensures all modalities for item $i$ point in consistent directions, yielding the aligned feature set  $\mathcal{Z}_i= \{\mathbf{z}_{i,1}, \mathbf{z}_{i,2}, \cdots, \mathbf{z}_{i,N_{m}} \}$, which provides a robust initialization for subsequent quantization.

\subsubsection{\textbf{Disentangled Modal-Residual Quantization (DMRQ)}}
\label{dmrq}
DMRQ discretizes the geometrically aligned embeddings via a ``coarse-to-fine'' strategy that structurally decouples shared cross-modal commonalities from modality-specific nuances, thereby preserving both consensus and characteristics while mitigating modality collapse.

Formally, given a user $u$ or an item $i$ with aligned multi-modal features $\mathcal{Z} = \{\mathbf{z}_{1}, \ldots, \mathbf{z}_{N_{m}}\}$, DMRQ maps $\mathcal{Z}$ to a discrete token sequence $\mathbf{c} = [\mathbf{c}_{sh}, \mathbf{c}_{sp}^{(1)}, \ldots, \mathbf{c}_{sp}^{(N_{m})}]$, where $\mathbf{c}_{sh}$ represents the shared consensus codes and $\mathbf{c}_{sp}^{(j)}$ captures the codes for modality $j$'s specific characteristics. The process begins by extracting the shared consensus through aggregating the aligned features into a global representation $\mathbf{f} = \Phi_{fuse}(\mathcal{Z})$. We then employ RQ-VAE~\citep{rajput2023recommender} to discretize $\mathbf{f}$ into $N_{sh}$ layers. Initializing $\mathbf{r}_0 = \mathbf{f}$, we recursively derive the code $c_{sh}^{(k)} = \arg\min_{v} \|\mathbf{r}_{k-1} - \mathbf{e}^{(k)}_v\|_2^2$ and update the residual $\mathbf{r}_k = \mathbf{r}_{k-1} - \mathbf{e}^{(k)}_{c_{sh}^{(k)}}$. The accumulated representation $\hat{\mathbf{z}}_{sh} = \sum_{k=1}^{N_{sh}} \mathbf{e}^{(k)}_{c_{sh}^{(k)}}$ captures the dominant cross-modal commonalities.


After shared quantization, the residual $\mathbf{r}_{N_{sh}}$ captures information not represented by the consensus codebook~\citep{zeghidour2021soundstream, lee2022autoregressive}. Through explicit disentanglement constraints (detailed below), we ensure that modality-specific details are preserved in this residual. To recover these characteristics for each modality, we introduce a \textbf{Parallel Semantically-Guided Recovery (PSGR)} mechanism. We first unfold $\mathbf{r}_{N_{sh}}$ into $H$ latent subspaces via multi-head projections to disentangle the features: $\tilde{\mathbf{K}}, \tilde{\mathbf{V}} \in \mathbb{R}^{H \times d_h}$. We then use the original aligned feature $\mathbf{z}_{j}$ as a semantic probe to selectively aggregate relevant subspaces: $\mathbf{z}_{sp}^{(j)} = \text{Attn}(\mathbf{z}_{j}, \tilde{\mathbf{K}}, \tilde{\mathbf{V}})$. The recovered continuous feature $\mathbf{z}_{sp}^{(j)}$ is then quantized to the nearest entry in the modality-specific codebook, yielding the code $\mathbf{c}_{sp}^{(j)}$ and its corresponding quantized vector $\hat{\mathbf{z}}_{sp}^{(j)}$.

To ensure the PSGR mechanism extracts purely modality-specific nuances, we impose a disentanglement constraint via Mutual Information (MI) minimization. This explicitly guides the attention to filter out redundant shared patterns and focus solely on characteristics statistically independent of the consensus $\hat{\mathbf{z}}_{sh}$. We employ the vCLUB estimator~\citep{cheng2020club} to optimize this (see Appendix~\ref{app:mi} for derivation): $\mathcal{L}_{MI} = \sum_{j=1}^{N_{m}} \hat{I}_{\text{vCLUB}}(\hat{\mathbf{z}}_{sh}; \mathbf{z}_{sp}^{(j)})$.


Finally, we optimize a unified objective that integrates compositional reconstruction with quantization and disentanglement constraints:
\begin{equation}
    \mathcal{L}_{DMRQ} = \sum_{j=1}^{N_{m}} \| \mathbf{z}_{j} - (\hat{\mathbf{z}}_{sh} + \hat{\mathbf{z}}_{sp}^{(j)}) \|_2^2 + \beta \mathcal{L}_{vq} + \lambda \mathcal{L}_{MI}
    \label{eq:dmrq_loss}
\end{equation}
The first term enforces an additive decomposition where the modality-specific component complements the shared base. The term $\mathcal{L}_{vq}$ aggregates the codebook commitment losses from both branches (detailed in Appendix~\ref{app:vq_loss}), while $\beta$ and $\lambda$ are hyperparameters balancing quantization stability and disentanglement.


\subsection{Hierarchical Memory-Anchor Transformer}

\label{sec:hmat}
We propose the \textbf{Hierarchical Memory-Anchor Transformer (HMAT)}, a specialized decoder-only architecture tailored for semantic ID-based recommendation. As illustrated in Figure~\ref{Hi-SAM}(b), HMAT adopts a stack of $N$ identical layers following the pre-normalization paradigm (utilizing RMSNorm and SwiGLU-based FFN), with the attention block fundamentally reconfigured via \textbf{H-RoPE} and \textbf{MA-Attn}. The state update rule for the $l$-th layer is:
\begin{align}
    \tilde{\mathbf{H}}^{(l)} &= \mathbf{H}^{(l-1)} + \text{MA-Attn}\left( \text{H-RoPE}(\mathbf{Q}^{(l-1)}, \mathbf{K}^{(l)}), \mathbf{V}^{(l)} \right) \\
    \mathbf{H}^{(l)} &= \tilde{\mathbf{H}}^{(l)} + \text{FFN}_{\text{SwiGLU}}\left(\text{RMSNorm}(\tilde{\mathbf{H}}^{(l)})\right)
\end{align}
where $\mathbf{Q}, \mathbf{K}, \mathbf{V}$ are projections of the normalized input. The two core modifications, H-RoPE and MA-Attn, are detailed below.


\textbf{Sequence Construction \& Coordinate Scheme.} 
We formulate the recommendation task as sequential transduction over a unified token stream $\mathcal{T}$. 
The user profile is represented as a sequence of tokens $\mathbf{c}_u$, and the $t$-th interacted item as $\mathbf{c}_t$.
To enable hierarchical information aggregation, we insert a special \textbf{Anchor Token} ($[\texttt{ANC}]$) after each item sequence but before the action token $a_t$. 
The global input sequence is constructed as:
\begin{equation}
    \mathcal{T} = [c_{u,1}, \dots, c_{u, L_u}, \dots, c_{t,1}, \dots, c_{t, L_i}, [\texttt{ANC}], a_t, \dots]
    \label{eq:seq_construction}
\end{equation}

To capture the intrinsic hierarchy of the stream—temporal evolution across items (\textit{Inter-Item}) and semantic composition within items (\textit{Intra-Item})—we assign a coordinate $(m, n)$ to each token. Here, $m$ denotes the global item order, and $n$ denotes the local attribute position, as illustrated in Figure~\ref{Hi-SAM}(b).
Formally, for tokens in the user profile, we set $m=0$; for tokens belonging to the $t$-th item (including its Anchor and Action), we set $m=t$. The intra-index $n$ resets to $1$ at the start of each new item segment.
In this layout, the Anchor Token serves as a semantic aggregator, compressing the fine-grained details of $\mathbf{c}_t$ into a holistic representation to predict the subsequent action.

\subsubsection{\textbf{Hierarchical Rotary Position Embedding (H-RoPE)}}
\label{sec:hrope}

Given the hierarchical coordinate $\left(m,n\right)$ defined above, we propose \textbf{H-RoPE} to inject the \textit{inter-item order} and \textit{intra-item position} into attention in a decoupled manner.
Concretely, we split the embedding dimension $d$ into two independent subspaces: the first $d/2$ dimensions encode the global item order $m$, and the remaining $d/2$ dimensions encode the local attribute position $n$.
For a token representation $\mathbf{x}\in\mathbb{R}^d$ at coordinate $\left(m,n\right)$, H-RoPE applies:
\begin{equation}
    \text{H-RoPE}(\mathbf{x}, m, n) = \left[ \mathcal{R}_{\text{inter}}(m) \mathbf{x}_{:d/2} \parallel \mathcal{R}_{\text{intra}}(n) \mathbf{x}_{d/2:} \right]
\end{equation}
where $\parallel$ denotes concatenation, and $\mathcal{R}_{\text{inter}}(m) = \text{diag}(\{e^{im\theta_j}\}_{j=1}^{d/4})$ applies rotation solely based on the item order $m$ (similarly for $\mathcal{R}_{\text{intra}}$).

To accommodate the asymmetric nature of recommendation sequences—where the inter-item history is extensive ($m$ is large, e.g., $>500$) while the intra-item composition is compact ($n$ is small, e.g., $\le 16$)—we assign distinct rotation base frequencies $\mathcal{B}$ to the two subspaces, defining the frequencies as $\theta_j = \mathcal{B}^{-2(j-1)/(d/2)}$.
Specifically, we set $\mathcal{B}_{\text{inter}} = 10^4$, which yields lower frequencies to ensure stable extrapolation over long histories, and $\mathcal{B}_{\text{intra}} = 100$ to induce higher frequencies that amplify sensitivity for local attributes.

\noindent \textbf{Decoupled Attention via H-RoPE.}
When H-RoPE is applied to both queries and keys, the attention score naturally decomposes along the two hierarchical dimensions.
For a query at $(m_q, n_q)$ and a key at $(m_k, n_k)$, the score decomposes into:
\begin{align}
    \label{eq:hrope_attn_main}
    & S_{\text{H-RoPE}}(\mathbf{q}, \mathbf{k}) = \text{Re} \left\langle \text{H-RoPE}(\mathbf{q}, m_q, n_q), \text{H-RoPE}(\mathbf{k}, m_k, n_k) \right\rangle \\
    &= \text{Re} \left( \langle \mathbf{q}_{\text{inter}}, \mathbf{k}_{\text{inter}}  e^{-i (\Delta m) \Theta_{\text{inter}}} \rangle + \langle \mathbf{q}_{\text{intra}}, \mathbf{k}_{\text{intra}}  e^{-i (\Delta n) \Theta_{\text{intra}}} \rangle \right)
\end{align}

where $\Delta m = m_q - m_k$ and $\Delta n = n_q - n_k$.
This shows that the two positional dimensions are strictly decoupled, with no cross-interference. Detailed derivation and the explicit expansion of Eq.~(\ref{eq:hrope_attn_main}) are provided in Appendix~\ref{app:hrope_derivation}.


\subsubsection{\textbf{Memory-Anchor Attention (MA-Attn)}}
\label{sec:maa}

To address the cumulative noise and computational inefficiency of modeling long, fine-grained semantic sequences, we propose \textbf{MA-Attn}.
Designed with the philosophy of human-like selective memory~\cite{richards2017persistence}—where historical events are retained only as compressed concepts—MA-Attn transforms the Anchor Token into a semantic condenser to filter out transient noise.

\noindent \textbf{Structured Attention Connectivity.}
To enforce this semantic compression, we restrict the attention topology based on the item index $m$. Let $m_q$ and $m_k$ denote the item indices of the query and key tokens, respectively. MA-Attn regulates information flow through three pathways:
(1) \textbf{Global User Context} ($m_k=0$): User profile tokens remain globally accessible to preserve invariant personalization.
(2) \textbf{Intra-Item Aggregation} ($m_q = m_k$): Tokens within the current item maintain full visibility to aggregate local attribute semantics into the Anchor.
(3) \textbf{Inter-Item Compressed Retrieval} ($m_k < m_q$): For historical items, access to raw tokens is blocked. Attention is routed exclusively to historical Anchor Tokens.

Formally, we inject this structural bias via a mask $\mathbf{M}$ into the attention mechanism:
\begin{equation}
    \text{MA-Attn}(\mathbf{Q}, \mathbf{K}, \mathbf{V}) = \text{Softmax}\left( S_{\text{H-RoPE}}(\mathbf{Q}, \mathbf{K}) + \mathbf{M} \right) \mathbf{V}
\end{equation}
where $S_{\text{H-RoPE}}(\mathbf{Q}, \mathbf{K})$ denotes the attention score matrix computed via H-RoPE (as defined in Eq.~\ref{eq:hrope_attn_main}), and $d$ is the head dimension. The attention bias $M_{q,k}$ is specifically defined as:
\begin{equation}
M_{q,k} =
\begin{cases}
0 & \text{if } m_k = 0 \lor m_q = m_k \\
0 & \text{if } m_k < m_q \land k = [\texttt{ANC}] \\
-\infty & \text{otherwise}
\end{cases}
\end{equation}
Note that causality ($k \le q$) is implicitly enforced. This design not only filters out historical noise but also renders raw tokens redundant, directly enabling the lossless cache eviction in Sec.~\ref{inference}.

\subsection{Training and Inference}

\subsubsection{\textbf{Training}}
Our framework is primarily optimized via Supervised Fine-tuning. To further enhance performance, we also introduce an optional progressive training strategy.
Throughout both stages, the DST remains frozen to maintain a stable discrete semantic space, decoupling representation stability from preference dynamics (see Appendix \ref{app:dlm}).

\noindent \textbf{Supervised Fine-tuning (SFT).}
The core optimization aligns the model with the recommendation task. 
In this stage, we \textbf{activate} the Memory-Anchor Mask $\mathbf{M}$ to restrict historical attention solely to Anchor Tokens.
We optimize the negative log-likelihood over the action tokens $\mathcal{I}_a$, conditioned on this sparsity-constrained context:
\begin{equation}
    \mathcal{L}_{\text{SFT}} = - \sum_{j \in \mathcal{I}_a} \log P(\mathcal{T}_j \mid \mathcal{T}_{<j}, \mathbf{M}; \Theta)
\end{equation}
where the dependency on $\mathbf{M}$ denotes that predictions rely on compressed memory states.

\noindent \textbf{Advanced Strategy: Semantic Pre-training.}
While SFT alone yields robust performance, we find that a preliminary pre-training stage can further improve convergence and semantic understanding.
Before SFT, we perform Next Token Prediction on the unified stream $\mathcal{T}$ with the Memory-Anchor Mask \textbf{disabled} (i.e., using a full causal mask).
This allows the model to learn the intrinsic co-occurrence patterns of semantic primitives by attending to the full context.
The objective is to minimize $\mathcal{L}_{\text{PT}} = - \sum_{j=1}^{|\mathcal{T}|} \log P(\mathcal{T}_j \mid \mathcal{T}_{<j}; \Theta)$ across the entire sequence.

\subsubsection{\textbf{Inference Optimization}}
\label{inference}
To enable high-throughput real-time recommendation, we implement a dual optimization strategy to minimize computational redundancy and memory bandwidth.

\noindent  \textbf{One-Pass Parallel Ranking.}  
Instead of evaluating candidates sequentially, we adopt the established One-Pass Parallel Ranking technique~\cite{han2025mtgr, xu2025climber} (e.g., aggregating candidates as $[\dots, \mathbf{c}_1, \dots, \mathbf{c}_k]$ with a block-diagonal mask) to compute scores in a single forward pass.
However, a naive flattening of candidates results in monotonically increasing position indices (e.g., $\mathbf{c}_k$ receives a much larger position ID than $\mathbf{c}_1$), introducing positional bias. To ensure ranking fairness, we implement Input-Side Position Re-alignment. By forcibly resetting the inter-item position coordinate $m$ of every candidate token to the effective history length $L_{valid} + 1$, we ensure that all candidates are evaluated under identical semantic contexts and positional embeddings, strictly independent of their batch order.


\noindent \textbf{Anchor-Based KV Cache Compression.}  
We leverage the structural sparsity of MA-Attn to implement strictly lossless \textbf{KV Cache Eviction}~\cite{zhang2023h2o}.
Since the mask $\mathbf{M}$ ensures that historical items are accessed exclusively via their Anchor Tokens, the fine-grained semantic tokens within those segments are never attended to by future tokens and become computationally redundant once their Anchor is generated.
We physically evict these redundant keys and values from the cache, retaining only the Anchor Tokens for history, and maintain a Logical Position Mapping to preserve the original coordinates for H-RoPE, ensuring correct relative position encoding despite the physical removal.
For a history of $K$ items with average length $L_i$, this reduces memory usage by $\sim L_i\times$ and attention complexity from $\mathcal{O}((K \cdot L_i)^2)$ to $\sim\mathcal{O}(K^2)$ for historical context.

\section{Experiments}
In this section, we evaluate Hi-SAM through extensive offline and online experiments on real-world industrial datasets, aiming to answer the following five research questions:

\noindent \textbf{RQ1:} How does Hi-SAM perform compared to state-of-the-art baselines in offline evaluation?

\noindent \textbf{RQ2:} How do different components and modalities contribute to the performance of Hi-SAM?

\noindent \textbf{RQ3:} Can Hi-SAM effectively align and disentangle multimodal semantics?

\noindent \textbf{RQ4:} Does Hi-SAM exhibit effective scaling behavior as computational resources increase?

\noindent \textbf{RQ5:} How does Hi-SAM perform in online industrial systems?

\begin{table}[t]
    \centering
    \caption{Overall statistics of the datasets. Avg. n denotes the average length of user interactions.}
    \label{tab:amazon_stats}
    \renewcommand{\arraystretch}{1} 
    \resizebox{1.0\columnwidth}{!}{ 
    \begin{tabular}{lrrrr}
        \toprule
        \textbf{Datasets} & \textbf{\#Users} & \textbf{\#Items} & \textbf{\#Inters.} & \textbf{Avg. n} \\
        \midrule
        Movies TV & 657.2K & 197.9K & 7.4M & 11.25  \\
        Book & 0.78M & 0.49M & 9.5M & 12.66   \\
        Industrial  & 6.3M & 1.38M & 521M & 82.30  \\
        \bottomrule
    \end{tabular}
    }
\end{table}

\begin{table*}[t]
    \centering
    \caption{Performance comparison on Public/Industrial Datasets. Best results are in \textbf{bold}, second-best are \underline{underlined}.}
    \label{tab:overall_performance}
    \renewcommand{\arraystretch}{1}
    \setlength{\tabcolsep}{3.5pt}
    \resizebox{\textwidth}{!}{
    \begin{tabular}{l|cccc|cccc|cccc}
        \toprule
        \multirow{2}{*}{\textbf{Method}} & \multicolumn{4}{c|}{\textbf{Book}} & \multicolumn{4}{c|}{\textbf{Movies \& TV}} & \multicolumn{4}{c}{\textbf{Industrial Dataset}} \\
        \cmidrule(lr){2-5} \cmidrule(lr){6-9} \cmidrule(lr){10-13}
         & \textbf{AUC} & \textbf{GAUC} & \textbf{Cold AUC} & \textbf{Cold GAUC} & \textbf{AUC} & \textbf{GAUC} & \textbf{Cold AUC} & \textbf{Cold GAUC} & \textbf{AUC} & \textbf{GAUC} & \textbf{Cold AUC} & \textbf{Cold GAUC} \\
        \midrule
        WuKong~\cite{zhang2024wukong} & 0.6910 & 0.6444 & 0.6878 & 0.6336 & 0.7494 & 0.7029 & 0.7586 & 0.7281 & 0.6266 & 0.6086 & 0.6709 & 0.5187 \\
        HSTU~\cite{zhai2024actions} & 0.6962 & 0.6440 & 0.6827 & 0.6544 & 0.7583 & 0.7076 & 0.7660 & 0.7359 & 0.6640 & 0.6087 & 0.6934 & 0.5304 \\
        MTGR~\cite{han2025mtgr} & 0.6967 & 0.6443 & 0.6831 & 0.6543 & 0.7601 & 0.7077 & 0.7667 & 0.7375 & 0.6812 & 0.6125 & 0.7103 & 0.5357 \\
        QARM~\cite{luo2025qarm}  &0.6964 & 0.6450 & 0.6872 & 0.6558 & 0.7616 & 0.7093 & 0.7673 & 0.7416 & 0.6420 & 0.6068 & 0.7467 & 0.5477 \\
        PSRQ+MCCA~\cite{wang2025progressive}  & 0.6969 & 0.6501 & 0.6877 & 0.6563 & 0.7622 & 0.7095 & 0.7696 & 0.7434 & 0.6803 & 0.6131 & 0.7524 & 0.5571 \\
        \midrule
        Hi-SAM-Small & 0.7060 & 0.6588 & 0.6924 & 0.6612 & 0.7816 & 0.7254 & 0.7861 & 0.7581 & 0.7293 & 0.6410 & 0.7886 & 0.5835 \\
        Hi-SAM-Large & \underline{0.7102} & \underline{0.6634} & \underline{0.6971} & \underline{0.6648} & \underline{0.7832} & \underline{0.7266} & \underline{0.7903} & \underline{0.7605} & \underline{0.7303} & \underline{0.6432} & \underline{0.7957} & \underline{0.5913} \\
        \quad \textit{w/} PT+SFT  & \textbf{0.7149} & \textbf{0.6660} & \textbf{0.6978} & \textbf{0.6661} & \textbf{0.7867} &  \textbf{0.7302} & \textbf{0.7943} & \textbf{0.7632} & \textbf{0.7337} & \textbf{0.6443} & \textbf{0.8028} & \textbf{0.5963} \\
        \bottomrule
    \end{tabular}
    }
\end{table*}

\subsection{Experimental Settings}

\noindent \textbf{Datasets.} We evaluate our method on one large-scale industrial dataset and two public benchmarks, with statistics in Table~\ref{tab:amazon_stats}. 
The industrial dataset is collected from a large online dating platform (September-December 2025), containing ~521M interactions among 6.33M users and 1.38M items. It includes chronological user behavior sequences (exposures, clicks, replies) with average length 82.30, along with multimodal information: avatar images, and textual content (personalized signatures, chat histories).  We also use the Movies TV and Books subsets from Amazon 2023~\cite{hou2024bridging} (May 1996-September 2023), extracting multimodal features including title, category, brand, and cover image. Following previous works~\cite{wang2020setrank, zhang2025collm}, ratings greater than 3 are treated as positive feedback and others as negative.For all datasets, interactions are chronologically sorted: the first 90\% for training and remaining 10\% for testing. Users with fewer than 10 interactions are defined as cold-start users for evaluation.


\noindent \textbf{Evaluation Metrics.} 
We employ AUC and GAUC as the primary offline metrics. AUC measures the overall ranking performance across all samples, while GAUC evaluates the intra-user ranking quality by averaging AUC over users. We report these metrics on both the entire test set and the cold-start subset to verify the model's effectiveness on general and sparse data distributions. For online evaluation, we conduct A/B testing focusing on Response Rate and Response Depth to assess the growth of business. 


\noindent \textbf{Baseline.} We compare Hi-SAM with 5 state-of-the-art sequential recommenders, including (1) 3 sparse ID-based recommenders: WuKong~\cite{zhang2024wukong}, HSTU~\cite{zhai2024actions}, and MTGR~\cite{han2025mtgr}; (2) 2 multimodal semantic ID-based recommenders: QARM~\cite{luo2025qarm} and PSRQ+MCCA~\cite{wang2025progressive}. 
To ensure a rigorous comparison, we strictly align both feature configurations and model complexity across all baselines. For input features, all methods utilize the same feature set, including interaction histories and item attributes. Notably, following its original design, MTGR additionally incorporates cross features (e.g., historical CTR) to enable interaction modeling in its generative framework. For model complexity, all baselines are configured with comparable computational costs: HSTU and MTGR use 4 transformer blocks, while WuKong, QARM, and PSRQ+MCCA are scaled accordingly.

\noindent \textbf{Implementation Details.} We instantiate Hi-SAM by configuring the DST and HMAT modules to integrate visual, textual, and behavioral modalities. For representation, we employ BLIP-2~\cite{li2023blip} (2.7B) and a SASRec-based encoder, projecting heterogeneous high-dimensional features into a unified 256-dimensional space. Through quantization, each item is encoded into 6 discrete semantic tokens ($N_{sh}=3$ for shared consensus, $N_{sp}=1$ per modality) with $H=8$ subspaces.  The HMAT module, incorporating MA-Attn and H-RoPE, is instantiated as a 4-layer architecture (hidden size 512) to align with baseline complexity, while a scaled-up 12-layer \textit{Hi-SAM Large} variant is evaluated to assess scalability. All models are trained on 8 NVIDIA A100 GPUs using the Adam optimizer. The maximum sequence length is standardized to 300 for all methods. Please refer to Appendix~\ref{app:implementation} for detailed implementation.

\subsection{Overall Performance (RQ1)}
Table~\ref{tab:overall_performance} presents the performance comparison across three datasets. Multimodal Semantic ID-based methods (QARM, PSRQ+MCCA) demonstrate clear advantages in cold-start scenarios. For instance, PSRQ+MCCA surpasses HSTU in Cold GAUC by approximately 5.0\% on the Industrial dataset (0.5571 vs. 0.5304) and 1.0\% on Movies \& TV (0.7434 vs. 0.7359), validating that multimodal semantic IDs effectively enhance generalization when interaction data is scarce. However, in overall evaluation, existing Semantic ID methods have not fully surpassed sparse ID-based counterparts. For example, QARM yields only a marginal 0.15\% GAUC gain over HSTU on the Book dataset (0.6450 vs. 0.6440), and even underperforms HSTU on the Industrial dataset (0.6068 vs. 0.6087), indicating that current tokenization and modeling approaches have not yet fully exploited multimodal information for overall ranking improvements, leaving considerable room for optimization.

Hi-SAM consistently outperforms all baselines across all metrics. On the Industrial dataset, Hi-SAM-Small improves GAUC from 0.6068 (QARM) and 0.6131 (PSRQ+MCCA) to 0.6410, and elevates Cold GAUC from 0.5477 and 0.5571 to 0.5835, respectively. This demonstrates that Hi-SAM's geometric alignment, modality-disentangled quantization, and hierarchical memory-anchor mechanism collectively enable more effective utilization of multimodal signals across both general and cold-start scenarios. Furthermore, scaling from Small to Large yields consistent gains (e.g., Cold GAUC from 0.5835 to 0.5913 on the Industrial dataset), demonstrating favorable scalability. The \textit{w/ PT+SFT} variant further pushes performance to state-of-the-art (Cold GAUC 0.5963), confirming that decoupling semantic learning from preference modeling is essential for maximizing the potential of multimodal recommendation.

\begin{table}[t]
    \centering
    \caption{Ablation study of decoupled modules: Tokenizers (Top) and Backbones (Bottom).}
    \label{tab:module_comparison}
    \renewcommand{\arraystretch}{1} 
    \resizebox{1.0\columnwidth}{!}{ 
    \begin{tabular}{l|cccc}
        \toprule
        \textbf{Module Variant} &  \textbf{AUC} & \textbf{GAUC} & \textbf{Cold AUC} & \textbf{Cold GAUC}  \\
        \midrule
        HSTU & 0.6640 & 0.6067 & 0.6934 & 0.5304 \\
        + QARM & 0.6622 & 0.6049 & 0.7292 & 0.5481 \\
        + PSRQ & 0.6703 & 0.6084 & 0.7443 & 0.5446 \\
        \textbf{+ DST (Ours)} & \textbf{0.7049} & \textbf{0.6244} & \textbf{0.7798} & \textbf{0.5795} \\
        \midrule
        QARM & 0.6420 & 0.6068 & 0.7467 & 0.5477 \\
        + HSTU Block & 0.6622 & 0.6049 &  0.7292 &   0.5481 \\
        + Qwen2.5 Block & 0.6909 & 0.6170 & 0.7549 & 0.5486 \\
        \textbf{+ HMAT (Ours)} & \textbf{0.7010} & \textbf{0.6270} & \textbf{0.7565} & \textbf{0.5573} \\
        \bottomrule
    \end{tabular}
    }
\end{table}

\begin{table}[t]
    \centering
    \caption{Ablation study of key components in Hi-SAM.}
    \label{tab:ablation}
    \renewcommand{\arraystretch}{1} 
    \resizebox{1.0\columnwidth}{!}{ 
    \begin{tabular}{l|cccc}
        \toprule
        \textbf{Variant} & \textbf{AUC} & \textbf{GAUC} & \textbf{Cold AUC} & \textbf{Cold GAUC}  \\
        \midrule
        \textbf{Hi-SAM (Full)} & \textbf{0.7293} & \textbf{0.6410} & \textbf{0.7886} & \textbf{0.5835}   \\
        \midrule
         w/o CGA & 0.6813 & 0.6166 & 0.7327 & 0.5465 \\
         w/o DMRQ & 0.7163 & 0.6347 & 0.7855 & 0.5822 \\
        \midrule
        w Abs. Pos. & 0.7241 & 0.6382 & 0.7824 & 0.5730 \\
        w RAB. Pos.  & 0.7247 & 0.6399 & 0.7832 & 0.5807 \\
        w 1D-Rope & 0.7260 & 0.6402 & 0.7850 & 0.5823 \\
        w/o MA-Attn  & 0.7201 & 0.6343 & 0.7845 & 0.5688 \\
        \bottomrule
    \end{tabular}
    }
\end{table}

\subsection{Ablation Study (RQ2)}
In this section, we conduct a systematic analysis on the industrial dataset to investigate the sources of Hi-SAM's performance improvements from three hierarchical levels: module-level, component-level, and modality-level.

\noindent  \textbf{Module-level Analysis.}
To ensure a fair comparison, we match key hyperparameters (e.g., tokenizer codebook size and decoder depth/width) across all variants to isolate structural differences. We first evaluate different tokenizers under the same HSTU backbone. As shown in Table \ref{tab:module_comparison}, incorporating multi-modal information consistently improves cold-start metrics over the ID-only baseline (e.g., PSRQ gains +7.34\% in Cold AUC). However, QARM and PSRQ yield suboptimal results: QARM suffers from modality collapse due to  early fusion, while PSRQ hinders cross-modal coupling due to independent quantization. In contrast, our DST achieves the strongest performance (e.g., +4.77\% Cold AUC over PSRQ) by effectively aligning modalities while preserving modality-specific details. Meanwhile, we compare transformer backbones using semantic IDs produced by QARM. The Qwen2.5 backbone outperforms HSTU when modeling semantic IDs. This is because HSTU's aggregation design tends to over-smooth the distinct semantic boundaries of quantized IDs, whereas Qwen2.5 utilizes softmax attention to precisely capture the deterministic dependencies among discrete tokens. Building on this, our HMAT backbone further improves GAUC over Qwen2.5 (+1.62\%) by incorporating position-aware and noise-filtering mechanisms.


\begin{figure}[t]
    \centering
    \includegraphics[width=1.0\linewidth]{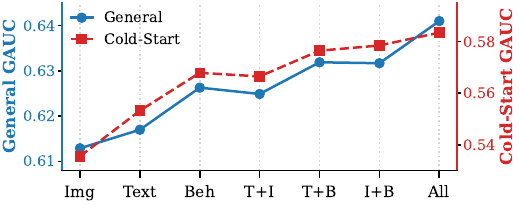}
    
    \caption{Ablation study on different modality combinations.}
    \label{fig:ablation}
    
\end{figure}

\begin{figure}[t]
  \centering
  \includegraphics[width=1.0\linewidth]{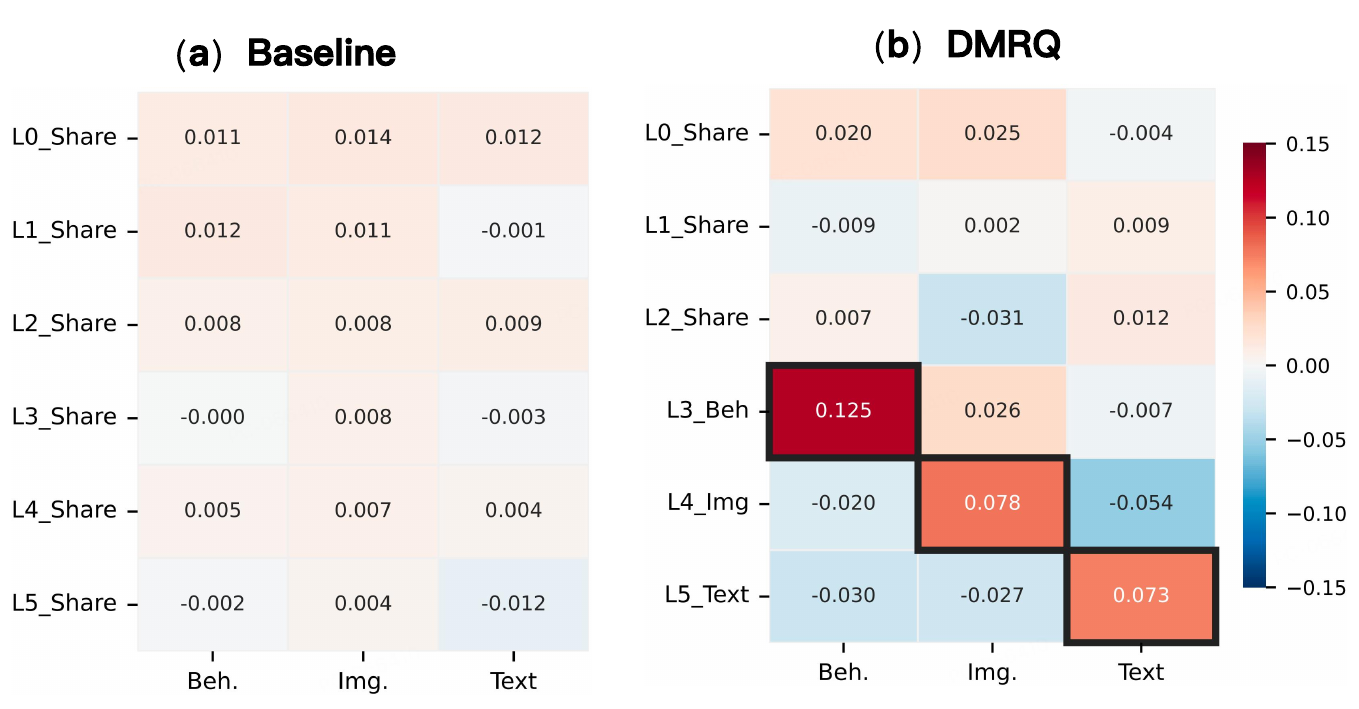}
  \caption{Layer-wise cosine similarity heatmaps of the baseline (a) and our DMRQ model (b) }
  \Description{Visualization of alignment and disentanglement}
  \label{fig:heatmap_only}
\end{figure}

\begin{figure*}[t]
    \centering
    \includegraphics[width=1.0\linewidth]{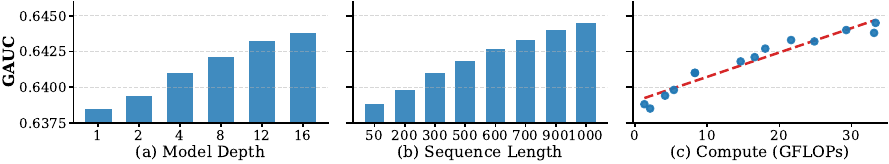}
    \caption{Scalability analysis of Hi-SAM regarding \textbf{(a)} model depth, \textbf{(b)} sequence length, and \textbf{(c)} computational cost (GFLOPs).}
    \label{fig:scalability}
\end{figure*}

\noindent  \textbf{Component-level Analysis.} 
Moving from macro-modules to micro-components, we investigate the necessity of specific technical designs within our framework, as detailed in Table \ref{tab:ablation}. Replacing CGA with naive concatenation (w/o CGA) results in a sharp performance decline of 3.81\% in GAUC, validating that effective cross-modal alignment is essential for our tokenization pipeline. Similarly, the degradation observed in w/o DMRQ (-0.98\% drop) underscores the necessity of our explicit separation strategy to better leverage fine-grained complementary information. Regarding the decoder, flat positional variants (e.g., 1D-RoPE) consistently underperform our H-RoPE (0.6402 vs. 0.6410 GAUC), highlighting the vital role of capturing the Item-Attribute hierarchy in user interactions. Finally, removing the Memory-Anchor Mask (w/o MA-Attn) causes a 1.05\% performance decline, confirming that our mechanism is beneficial for semantic ID-based recommendation.

\noindent  \textbf{Modality-level Contribution Analysis.} 
To quantify the contribution of each modality, we conduct an ablation study by evaluating different modality combinations, as visualized in Figure~\ref{fig:ablation}.  First, \textit{Behavior-only} serves as a strong baseline, significantly outperforming \textit{Text-only} and \textit{Image-only} variants. Notably, it even surpasses the \textit{Text+Image} combination, confirming that collaborative signals from user interactions remain the primary source for preference modeling.
Second, coupling behavior with either text or image consistently outperforms the single-modality baseline, particularly in cold-start scenarios (e.g., \textit{Image+Behavior} improves Cold AUC by +3.50\%), indicating that semantic cues effectively compensate for sparse interactions. Most importantly, the full tri-modal Hi-SAM achieves the highest performance across all metrics (AUC +3.05\% over \textit{Image+Behavior}), suggesting that behavioral, textual, and visual modalities provide effective complementary information within our framework.

\subsection{Visualization of Modal Disentanglement and Alignment (RQ3)}
We analyze the internal mechanisms of our Disentangled Semantic Tokenizer by visualizing the layer-wise code correlations and the topological structure of the latent space. We first validate the effectiveness of the disentanglement design in DMRQ via layer-wise similarity heatmaps. As shown in Figure~\ref{fig:heatmap_only}, the baseline w/o PSGR (Left) displays a relatively uniform distribution, indicating that multi-modal details remain globally entangled within the residuals. In contrast, our DMRQ (Right) reveals a clear “Coarse-to-Fine” hierarchy. The first three layers ($L_0 \sim L_2$) exhibit balanced correlations similar to the baseline, confirming they encode the shared consensus. However, a sharp diagonal pattern emerges in deeper layers, where $L_3$, $L_4$, and $L_5$ correlate strongly with Behavior, Image, and Text, respectively. This confirms that our PSGR mechanism successfully retrieves specific modal nuances from mixed residuals and routes them into dedicated subspaces. Additional 3D t-SNE plots in Appendix~\ref{sec:appendix_tsne} further demonstrate that our model successfully aligns multi-modal embeddings into structured clusters, significantly improving over the chaotic raw space.

\subsection{ Scalability (RQ4)}
We examine the scalability of Hi-SAM by varying model depth ($L$) and sequence length ($S$). The computational cost is measured in GFLOPs, which scales quadratically with sequence length and linearly with model depth (i.e., $\text{GFLOPs} \propto L \cdot S^2$). Figure~\ref{fig:scalability}(c) demonstrates a favorable scaling law: GAUC increases consistently with computational investment, indicating predictable performance gains. As detailed in Figure~\ref{fig:scalability}(a) and (b), the model establishes a robust baseline (GAUC > 0.638) even at minimal settings (e.g., $L=1$ or $S=50$). Starting from this foundation, increasing depth from 1 to 16 layers (with $S=300$) improves GAUC to 0.6438, while extending sequence length from 50 to 1000 (with $L=4$) raises it to 0.6445. Both dimensions exhibit power-law-like scaling, characterized by rapid initial gains that gradually saturate, confirming that expanding model capacity and context effectively translates to higher accuracy.

\begin{table}[t]
    \centering
    \caption{Online A/B testing: Hi-SAM variants vs. baseline.}
    \label{tab:online_ab}
    \renewcommand{\arraystretch}{1.1}
    \resizebox{\columnwidth}{!}{
    \begin{tabular}{l|cc|cc}
        \toprule
        \multirow{2}{*}{\textbf{Model Variant}} & \multicolumn{2}{c|}{\textbf{ALL }} & \multicolumn{2}{c}{\textbf{Cold-Start }} \\
        \cmidrule(lr){2-3} \cmidrule(lr){4-5}
         & \textbf{Resp. Rate} & \textbf{Resp. Depth} & \textbf{Resp. Rate} & \textbf{Resp. Depth} \\
        \midrule
        Hi-SAM Large (L=200) & +2.31\% & -0.77\% & +13.58\% & +0.93\% \\
        Hi-SAM Large (L=400) & +3.71\% & +1.86\% & +13.68\% & +5.74\% \\
        \quad \textit{w/} PT+SFT & \textbf{+6.55\%} & \textbf{+5.48\%} & \textbf{+16.62\%} & \textbf{+8.91\%} \\
        \bottomrule
    \end{tabular}
    }
\end{table}

\subsection{Online Experiments (RQ5)}
To rigorously validate Hi-SAM, we conducted A/B testing on 6\% of live traffic over a two-month period on a large-scale social platform. The model has since been deployed in production serving millions of daily active users. The experiment benchmarks against a highly optimized DLRM with years of continuous online iteration.
Table~\ref{tab:online_ab} reports the relative improvements over the baseline. The Hi-SAM Large (L=200) yields a 2.31\% gain in Response Rate but shows a slight decrease of 0.77\% in Response Depth. Extending the sequence length to 400 addresses this, achieving positive gains across both metrics (+3.71\% and +1.86\%, respectively). The \textit{PT+SFT} strategy further boosts performance, achieving +6.55\% in Response Rate and +5.48\% in Response Depth. Notably, for cold-start users, the final variant achieves a +16.62\% lift in Response Rate, demonstrating strong robustness when interaction history is sparse.  In the online inference stage, Hi-SAM achieves a 35\% reduction in Response Time compared to the baseline under the same computational budget, enabled by our optimization strategies (Section~\ref{inference}). This efficiency gain allows us to deploy a multimodal model with significantly higher complexity than DLRM within strict latency constraints.



\section{Conclusion}
We propose Hi-SAM, a hierarchical structure-aware multi-modal framework for semantic ID-based recommendation. Hi-SAM introduces a Disentangled Semantic Tokenizer that combines geometric alignment with disentangled quantization to preserve both cross-modal consensus and modality-specific nuances, and a Hierarchical Memory-Anchor Transformer that explicitly models the hierarchical data structure through decoupled positional encoding and anchor-based sequence compression. Extensive offline experiments demonstrate consistent improvements over state-of-the-art baselines, especially in cold-start scenarios. Online A/B testing further validates its effectiveness with a 6.55\% Response Rate gain and 35\% latency reduction. Hi-SAM has been fully deployed on a large-scale social platform serving millions of daily active users.

\bibliographystyle{ACM-Reference-Format}
\balance
\bibliography{arxiv}

\appendix

\section{Supplement to Method}
\label{sec:appendix}

\subsection{Derivation of Mutual Information Minimization}
\label{app:mi}

In this section, we provide the detailed derivation of the Mutual Information (MI) minimization constraint used in the DMRQ module. Our objective is to explicitly disentangle the shared consensus representation $\hat{\mathbf{z}}_{sh}$ from the modality-specific recovered features $\mathbf{z}_{sp}^{(j)}$. Mathematically, this is achieved by minimizing the Mutual Information $I(\hat{\mathbf{z}}_{sh}; \mathbf{z}_{sp}^{(j)})$. Since the true joint distribution is unknown and high-dimensional, direct computation is intractable.

To address this, we employ the \textbf{Variational Contrastive Log-ratio Upper Bound (vCLUB)}~\citep{cheng2020club}. It is crucial to note that while lower bounds (like InfoNCE) are suitable for maximizing MI, disentanglement requires minimizing an \textit{upper bound} to effectively reduce the correlation. vCLUB utilizes a variational distribution $q_{\theta}(\mathbf{z}_{sp}^{(j)} \mid \hat{\mathbf{z}}_{sh})$, parameterized by a neural network, to approximate the true conditional distribution $p(\mathbf{z}_{sp}^{(j)} \mid \hat{\mathbf{z}}_{sh})$. The upper bound is derived based on the non-negativity of KL-divergence:
\begin{equation}
\begin{split}
    I(\hat{\mathbf{z}}_{sh}; \mathbf{z}_{sp}^{(j)}) \leq \;& \mathbb{E}_{p(\hat{\mathbf{z}}_{sh}, \mathbf{z}_{sp}^{(j)})} [\log q_{\theta}(\mathbf{z}_{sp}^{(j)} \mid \hat{\mathbf{z}}_{sh})] \\
    &- \mathbb{E}_{p(\hat{\mathbf{z}}_{sh})p(\mathbf{z}_{sp}^{(j)})} [\log q_{\theta}(\mathbf{z}_{sp}^{(j)} \mid \hat{\mathbf{z}}_{sh})]
\end{split}
\end{equation}

In our implementation, we model the variational approximation $q_{\theta}$ as a Gaussian distribution $\mathcal{N}(\mu_{\theta}(\hat{\mathbf{z}}_{sh}), \sigma^2_{\theta}(\hat{\mathbf{z}}_{sh})\mathbf{I})$, where $\mu_{\theta}$ and $\sigma_{\theta}$ are inferred by a MLP. Given a mini-batch of $B$ samples, the unbiased estimator $\hat{I}_{\text{vCLUB}}$ is calculated as:
\begin{equation}
    \hat{I}_{\text{vCLUB}} = \frac{1}{B} \sum_{k=1}^{B} \log q_{\theta}(\mathbf{z}_{sp}^{(j, k)} \mid \hat{\mathbf{z}}_{sh}^{(k)}) - \frac{1}{B^2} \sum_{k=1}^{B} \sum_{l=1}^{B} \log q_{\theta}(\mathbf{z}_{sp}^{(j, l)} \mid \hat{\mathbf{z}}_{sh}^{(k)})
\end{equation}
The first term represents the log-likelihood of positive pairs (from the joint distribution), while the second term averages over all possible pairs in the batch to approximate the product of marginals. During training, we alternately update the variational approximator $q_{\theta}$ to maximize the log-likelihood (ensuring accurate estimation) and the encoder parameters to minimize $\hat{I}_{\text{vCLUB}}$ (achieving disentanglement).

\subsection{Details of Quantization Objective}
\label{app:vq_loss}

The quantization loss term $\mathcal{L}_{vq}$ in Eq.~\ref{eq:dmrq_loss} stabilizes codebook learning by pulling codebook vectors toward encoder outputs (codebook loss) and preventing encoder outputs from drifting (commitment loss), following the RQ-VAE paradigm~\cite{rajput2023recommender}.

Since DMRQ involves a hierarchical quantization process (Shared + Specific), $\mathcal{L}_{vq}$ is composed of two parts:
\begin{equation}
    \mathcal{L}_{vq} = \mathcal{L}_{vq}^{sh} + \mathcal{L}_{vq}^{sp}
\end{equation}

For the Shared Branch, which employs Residual Quantization with depth $N_{sh}$, the loss is accumulated across all residual steps:
\begin{equation}
    \mathcal{L}_{vq}^{sh} = \sum_{k=1}^{N_{sh}} \left( \| \text{sg}[\mathbf{r}_{k-1}] - \mathbf{e}^{(k)}_{c_{sh}^{(k)}} \|_2^2 + \gamma \| \mathbf{r}_{k-1} - \text{sg}[\mathbf{e}^{(k)}_{c_{sh}^{(k)}}] \|_2^2 \right)
\end{equation}
where $\text{sg}[\cdot]$ denotes the stop-gradient operator, $\mathbf{r}_{k-1}$ is the input residual to layer $k$, and $\mathbf{e}^{(k)}$ is the selected codebook vector.

For the Specific Branch, the loss is applied to the recovered feature $\mathbf{z}_{sp}^{(j)}$ for each modality $j$:
\begin{equation}
    \mathcal{L}_{vq}^{sp} = \sum_{j=1}^{N_m} \left( \| \text{sg}[\mathbf{z}_{sp}^{(j)}] - \hat{\mathbf{z}}_{sp}^{(j)} \|_2^2 + \gamma \| \mathbf{z}_{sp}^{(j)} - \text{sg}[\hat{\mathbf{z}}_{sp}^{(j)}] \|_2^2 \right)
\end{equation}
Here, $\gamma$ is the commitment coefficient, set to $0.25$ in our experiments. This formulation ensures that both the shared consensus and the specific nuances are mapped to their respective discrete spaces with high fidelity.

\subsection{Detailed Derivation of H-RoPE}
\label{app:hrope_derivation}

In this section, we provide the detailed derivation of the attention score presented in Eq.~(\ref{eq:hrope_attn_main}).
To facilitate the derivation, we first reformulate the vector-valued function $\text{H-RoPE}(\mathbf{x}, m, n)$ in the complex domain.
Given a vector $\mathbf{x} \in \mathbb{R}^d$ and the split dimension $d$, the complex representation is:
\begin{equation}
\begin{split}
    &\text{H-RoPE}(\mathbf{x}, m, n) \cong \\
    &\begin{pmatrix}
    (x_0 + i x_1) e^{i m \theta_{\text{inter}, 0}} \\
    \vdots \\
    (x_{d/2-2} + i x_{d/2-1}) e^{i m \theta_{\text{inter}, d/4-1}} \\
    (x_{d/2} + i x_{d/2+1}) e^{i n \theta_{\text{intra}, 0}} \\
    \vdots \\
    (x_{d-2} + i x_{d-1}) e^{i n \theta_{\text{intra}, d/4-1}}
    \end{pmatrix}
\end{split}
\end{equation}
Substituting this complex form into the inner product of Eq.~(\ref{eq:hrope_attn_main}), we obtain the expanded attention score \begin{equation}
\label{eq:detailed_score}
\begin{split}
    &S_{\text{H-RoPE}}(\mathbf{q}, \mathbf{k}) \\
    &= \text{Re} \left\langle \text{H-RoPE}(\mathbf{q}, m_q, n_q), \text{H-RoPE}(\mathbf{k}, m_k, n_k) \right\rangle \\
    &= \sum_{j=0}^{d/4-1} \Big[ (q_{2j} k_{2j} + q_{2j+1} k_{2j+1}) \cos((m_q - m_k)\theta_{\text{inter}, j}) \\
    &\quad\quad + (q_{2j} k_{2j+1} - q_{2j+1} k_{2j}) \sin((m_q - m_k)\theta_{\text{inter}, j}) \Big] \\
    &+ \sum_{j=0}^{d/4-1} \Big[ (q_{d/2+2j} k_{d/2+2j} + q_{d/2+2j+1} k_{d/2+2j+1}) \\
    &\quad\quad\quad \cdot \cos((n_q - n_k)\theta_{\text{intra}, j}) \\
    &\quad\quad + (q_{d/2+2j} k_{d/2+2j+1} - q_{d/2+2j+1} k_{d/2+2j}) \\
    &\quad\quad\quad \cdot \sin((n_q - n_k)\theta_{\text{intra}, j}) \Big]
\end{split}
\end{equation}
As shown in Eq.~(\ref{eq:detailed_score}), the attention score naturally decomposes into two independent terms governed by relative distances $\Delta m = m_q - m_k$ and $\Delta n = n_q - n_k$ respectively.
This explicit expansion verifies the decoupled nature of H-RoPE as claimed in Section~\ref{sec:hrope}.

\subsection{Decoupled Lifecycle Management}
\label{app:dlm}

We implement a multi-tiered lifecycle management strategy based on the stability of different model components.
First, regarding semantic representation, item semantics captured by the DST are relatively stable. Therefore, we keep the tokenizer frozen during downstream training and only update it at a low frequency (e.g., monthly) to adapt to long-term data distribution changes. This prevents "Semantic Shifts" and ensures a stable feature space.
Second, for user preference modeling, we differentiate between general semantic understanding and task-specific alignment.
The Semantic Pre-training (PT) stage, which learns general sequence dependencies, is updated with medium frequency (e.g., weekly) to maintain robust convergence and understanding.
In contrast, the Supervised Fine-tuning (SFT) stage is updated with high frequency (e.g., daily) to capture real-time shifts in user interests.
This hierarchical decoupling ensures the model remains both robust to evolving content and responsive to immediate user behaviors.

\section{Experimental Settings}

\subsection{Baselines}
\label{app:baselines}

We evaluate Hi-SAM against two groups of state-of-the-art baseline methods:

\noindent \textbf{(1) Sparse ID-based Recommenders}: These methods primarily rely on sparse features and ID sequences, representing the current industrial standard for large-scale retrieval and ranking.
\begin{itemize}[leftmargin=*]
    \item \textbf{WuKong} \cite{zhang2024wukong} proposes a network architecture based on \textit{stacked factorization machines} to establish scaling laws in recommendation. It captures diverse, any-order interactions through deeper and wider layers to handle complex real-world datasets.
    \item \textbf{HSTU} \cite{zhai2024actions} reformulates recommendation as a sequential transduction task within a \textit{Generative Recommender} framework. It introduces a high-performance architecture designed for high-cardinality, non-stationary data, demonstrating that model quality scales as a power-law of training compute.
    \item \textbf{MTGR} \cite{han2025mtgr} addresses the performance degradation in generative models caused by abandoning traditional cross features. Built upon the HSTU architecture, it integrates \textit{cross features} (e.g., historical CTR) and employs Group-Layer Normalization to enable efficient industrial-scale generative recommendation.
\end{itemize}

\noindent \textbf{(2) Multimodal Semantic ID-based Recommenders}: These methods utilize quantization techniques to incorporate multimodal semantics into discrete tokens for unified modeling.
\begin{itemize}[leftmargin=*]
    \item \textbf{QARM} \cite{luo2025qarm} addresses the "representation unmatching" and "unlearning" issues in multimodal recommendation. It employs an item alignment module to match user behavior distributions and generates \textit{trainable quantitative codes} to adapt pre-trained representations for downstream ranking tasks.
    \item \textbf{PSRQ+MCCA} \cite{wang2025progressive} proposes a two-stage framework for music recommendation. It utilizes Progressive Semantic Residual Quantization (PSRQ) to preserve prefix semantics during discretization, and a Multi-Codebook Cross-Attention (MCCA) network to simultaneously capture modal-specific interests and cross-modal correlations.
\end{itemize}

\begin{figure}[t]
  \centering
  \includegraphics[width=1.0\linewidth]{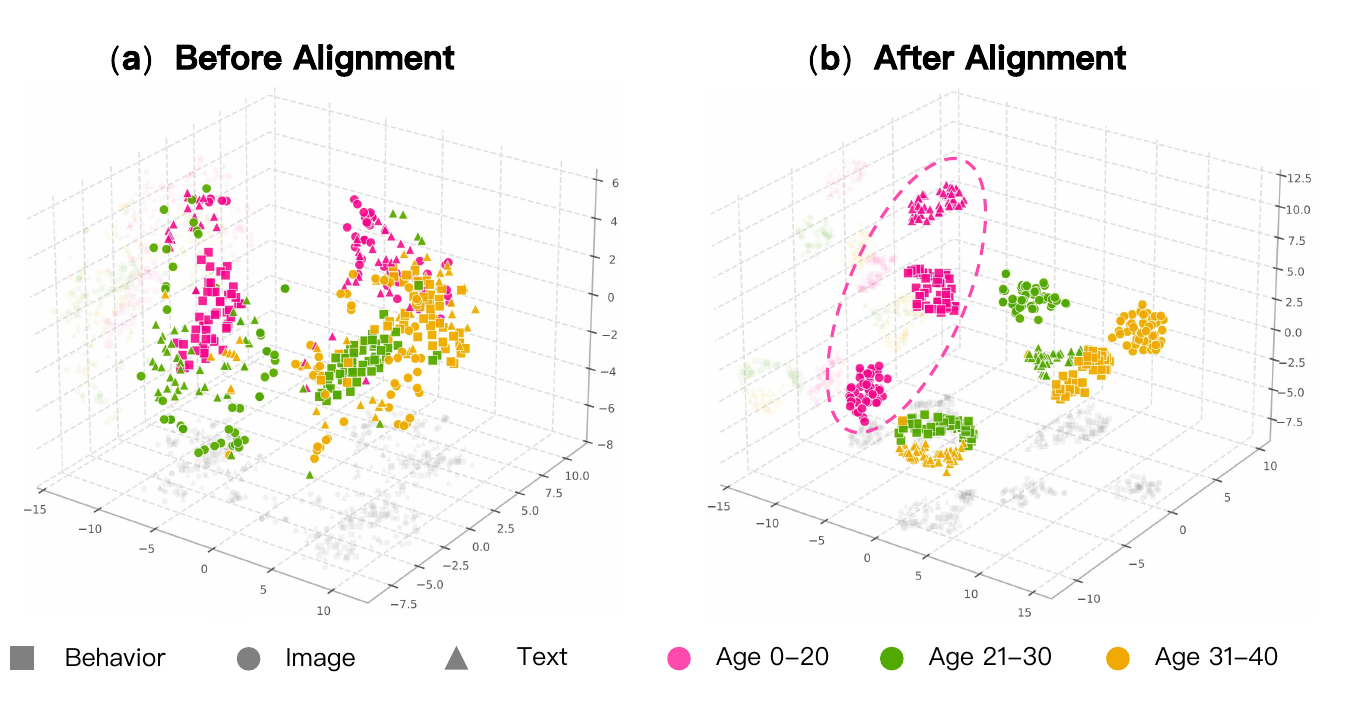}
  \caption{3D t-SNE visualizations of the latent space before (a) and after (b) alignment. }
  \Description{Visualization of alignment }
  \label{fig:appendix_tsne}
\end{figure}

\subsection{Implementation Details}
\label{app:implementation}
We instantiate Hi-SAM by configuring the DST and HMAT modules to integrate three distinct modalities: visual, textual, and behavioral signals. Specifically, the tokenizer employs BLIP-2~\cite{li2023blip} (2.7B) for visual ($d=2560$) and textual ($d=1408$) features, alongside a SASRec-based encoder for behavioral embeddings ($d=512$). To prevent data leakage, the SASRec encoder is trained on samples strictly isolated by time from the downstream ranking data. These heterogeneous features are aligned into a unified 256-dimensional space via CGA. For quantization, we configure $N_{sh}=3$ shared codebooks to capture consensus and assign $N_{sp}=1$ specific codebook per modality to preserve nuances, resulting in a total of 6 codebooks (codebook size $512 \times 256$). Specifically, within the PSGR module, we set the number of subspaces $H=4$. For fair comparison, the total number of codebooks in baseline methods is maintained consistent with ours. Subsequently, the decoder backbone is configured with a hidden size of 512 and FFN size of 2560. To optimize inference efficiency, we implement MA-Attn using Grouped Query Attention (GQA) with 8 query heads and 2 key-value heads. The HMAT depth is set to 4 layers to align with baseline complexity, while the industrial \textit{Hi-SAM Large} variant is scaled to 12 layers. Additionally, H-RoPE base frequencies are set to 10,000 (inter-item) and 100 (intra-item) to enhance local positional sensitivity. The framework is implemented using Python 3.11.9 and PyTorch 2.4.1, utilizing DeepSpeed ZeRO-2 and FP16 for efficiency. We optimize the model via Adam (batch size 128) on 8 NVIDIA A100 GPUs, with learning rates of $2 \times 10^{-4}$ for pre-training and $1 \times 10^{-4}$ for SFT. Regarding baselines, we use the official implementations for WuKong and HSTU, and strictly follow the original papers for MTGR, QARM, and PSRQ+MCCA. The maximum sequence length is standardized to 300 for all methods.

\section{More Experimental Results}
\subsection{Additional Visualization of Latent Space}
\label{sec:appendix_tsne}
To further analyze the Cross-Modal Alignment, we visualize user embeddings stratified by age groups (0-20, 21-30, 31-40) using 3D t-SNE. Figure~\ref{fig:appendix_tsne} presents the comparison between the raw feature space and the aligned space learned by our model. 

As shown in Figure~\ref{fig:appendix_tsne} (a) (Before Alignment), the raw space exhibits a chaotic distribution where age groups and modalities are inextricably mixed, indicating a significant modality gap. In contrast, Figure~\ref{fig:appendix_tsne} (b) (After Alignment) demonstrates that our model generates a structured space with distinct age clusters. Within these clusters (e.g., the dashed circle), embeddings from Image, Text, and Behavior are tightly aligned according to a consistent topology. This confirms that our Disentangled Semantic Tokenizer successfully bridges the modality gap while preserving user-specific semantics.

\end{document}